\documentclass{article} 
\usepackage{arxiv-style/arxiv}
\usepackage{booktabs}
\usepackage{amsmath, amsfonts, amssymb}
\usepackage[hidelinks]{hyperref} 
\usepackage{cleveref}
\usepackage{subcaption}

\newcommand{\vect}[1]{\boldsymbol{\mathbf{#1}}}
\newcommand{\hp}[1]{\texttt{#1}}
\date{}

\usepackage[disable]{todonotes}
\newcommand{\romain}[1]{\todo[inline,color=purple!40]{#1 -- Romain}} 

\title{Streamlining Ocean Dynamics Modeling with Fourier Neural Operators: A Multiobjective Hyperparameter and Architecture Optimization Approach}
\newif\ifuniqueAffiliation
\uniqueAffiliationtrue

\ifuniqueAffiliation % Standard variant of author block
\author{ \href{https://orcid.org/0000-0003-1109-3380}{\includegraphics[scale=0.06]{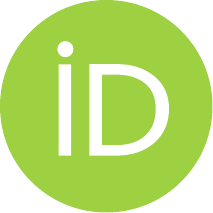}\hspace{1mm}Yixuan Sun}\\
	Argonne National Laboratory\\
	\texttt{yixuan.sun@anl.gov} \\
	\And
	Ololade Sowunmi\\
	Florida State University\\
	\texttt{osowunmi@fsu.edu} \\
        \And
        \href{https://orcid.org/0000-0002-8992-8192}{\includegraphics[scale=0.06]{orcid.pdf}\hspace{1mm}Romain Egele}\\
        Université Paris-Saclay\\
        \texttt{romain.egele@universite-paris-saclay.fr}\\
        \And
        \href{https://orcid.org/0000-0003-0388-5943}{\includegraphics[scale=0.06]{orcid.pdf}\hspace{1mm}Sri Hari Krishna Narayanan}\\
	Argonne National Laboratory\\
	\texttt{snarayan@mcs.anl.gov} \\
        \And
        \href{https://orcid.org/0000-0003-1418-5686}{\includegraphics[scale=0.06]{orcid.pdf}\hspace{1mm}Luke Van Roekel}\\
        Los Alamos National Laboratory\\
        \texttt{lvanroekel@lanl.gov}
        \And
        \href{https://orcid.org/0000-0002-0292-5715}{\includegraphics[scale=0.06]{orcid.pdf}\hspace{1mm}Prasanna Balaprakash\thanks{Corresponding Author}}\\
        Oak Ridge National Laboratory\\
        \texttt{pbalapra@ornl.gov}
}
\else
\usepackage{authblk}

\newbox{\orcid}\sbox{\orcid}{\includegraphics[scale=0.06]{orcid.pdf}} 
\author[1]{%
	\href{https://orcid.org/0000-0000-0000-0000}{\usebox{\orcid}\hspace{1mm}David S.~Hippocampus\thanks{\texttt{hippo@cs.cranberry-lemon.edu}}}%
}
\author[1,2]{%
	\href{https://orcid.org/0000-0000-0000-0000}{\usebox{\orcid}\hspace{1mm}Elias D.~Striatum\thanks{\texttt{stariate@ee.mount-sheikh.edu}}}%
}
\affil[1]{Department of Computer Science, Cranberry-Lemon University, Pittsburgh, PA 15213}
\affil[2]{Department of Electrical Engineering, Mount-Sheikh University, Santa Narimana, Levand}
\fi

\renewcommand{\shorttitle}{}

\hypersetup{
pdftitle={A template for the arxiv style},
pdfsubject={q-bio.NC, q-bio.QM},
pdfauthor={David S.~Hippocampus, Elias D.~Striatum},
pdfkeywords={First keyword, Second keyword, More},
}

\definecolor{grayblue}{RGB}{112,146,190} % This is just an example color

\setuptodonotes{color=grayblue!30}
\begin{document}
\maketitle
\begin{abstract}
    
Training an effective deep learning model to learn ocean processes involves careful choices of various hyperparameters. We leverage the advanced search algorithms for multiobjective optimization in DeepHyper, a scalable hyperparameter optimization software, to streamline the development of neural networks tailored for ocean modeling. The focus is on optimizing Fourier neural operators (FNOs), a data-driven model capable of simulating complex ocean behaviors. Selecting the correct model and tuning the hyperparameters are challenging tasks, requiring much effort to ensure model accuracy. DeepHyper allows efficient exploration of hyperparameters associated with data preprocessing, FNO architecture-related hyperparameters, and various model training strategies. We aim to obtain an optimal set of hyperparameters leading to the most performant model. Moreover, on top of the commonly used mean squared error for model training, we propose adopting the negative anomaly correlation coefficient as the additional loss term to improve model performance and investigate the potential trade-off between the two terms. The experimental results show that the optimal set of hyperparameters enhanced model performance in single timestepping forecasting and greatly exceeded the baseline configuration in the autoregressive rollout for long-horizon forecasting up to 30 days. Utilizing DeepHyper, we demonstrate an approach to enhance the use of FNOs in ocean dynamics forecasting, offering a scalable solution with improved precision.
\end{abstract}

\listoftodos

\section{Introduction}
Deep learning techniques significantly enhance scientific computing, serving as efficient surrogate models for complex, time-intensive first-principles-driven simulations. Their application in climate modeling leverages large amounts of high-resolution, multidimensional data. It addresses the computational challenges posed by increasingly complex climate systems, resulting in remarkable advances in environmental research~\cite{kurth2018exascale, rasp2018deep, nguyen2023climax, gibson2021training, pathak2022fourcastnet}. 

%Concurrently, the critical role of ocean processes within the climate system is gaining increased focus. 
The oceans play a leading role in the climate system via the storage and transport of anthropogenically generated heat and carbon dioxide, a role that is crucial for mitigating climate change. Simultaneously, oceanic processes that redistribute mass, heat, and salt are instrumental in driving phenomena such as hurricanes, extreme precipitation, and droughts~\cite{cheng2017improved}. Deep learning models have empowered ocean modeling from a spatial-temporal forecast of select ocean features~\cite{gou2020deepocean, choi2022improving, partee2022using} to parameterization~\cite{zhu2022physics, guillaumin2021stochastic, zanna2020data}. Despite the success of those models, however, selecting and fine-tuning appropriate model hyperparameters remain challenging, often requiring a significant amount of manual search via trial and error. 

Hyperparameters are the non-trainable parameters determining the data preprocessing, network structure, and training procedures; and they highly impact the performance of deep learning models~\cite{liao2022empirical}. Searching for a proper combination that works optimally for the specific problem can be challenging. With the increasing complexity of performant deep learning models,  manual tuning conducted in a trial-and-error way becomes infeasible, given that hyperparameter configurations exist within a vast search space. 

Deep learning models for ocean processes must accurately capture the mean behavior of the ocean state profile and account for the detailed variation throughout the domain. In modeling climate processes using deep learning models, one typically formulates the problem as an image-to-image regression problem, where the pixels of an image represent the state values within the domain of interest and the image channels are associated with different state variables. One often uses per-pixel mean squared error~(MSE) as the loss function because of differentiability and correspondence to maximized likelihood with assumed Gaussian errors to train these models~\cite{pathak2022fourcastnet, bi2022panguweather, nguyen2023climax}. However, MSE penalizes large errors more significantly and usually produces images in favor of averaged pixel values that may be blurry and unnatural~\cite{mustafa2021training, zhao2016loss}. This situation might become problematic in producing ocean dynamics forecasts because the values are relatively stable and have a long time scale compared with the weather, where the values evolve closely to the mean field.
% \todo{Check the statement}. 
Many modifications and different loss functions have been proposed to mitigate the issue of MSE~\cite{johnson2016perceptual, mustafa2021training}, but they are highly problem-dependent and can be complex. Here, specifically for ocean modeling with deep neural networks, we propose using the negative value of a simple metric, the anomaly correlation withefficient~(ACC)~\cite{murphy1989skill}, commonly used in climate model evaluation, as the additional loss term on top of the MSE to address this potential issue.   

% The trade-off between the model expressivity, usually associated with high performance, and model complexity, corresponding to longer inference time, must be skillfully balanced through the optimal combination of the hyperparameters. In this study, we aim to streamline ocean modeling with deep neural networks, searching for the optimal hyperparameter configurations that lead to accurate one-step forward forecasting, which could improve mid to long-range autoregressive forecasting.

% Deep learning powered climate modeling. Scientific machine learning frameworks for FNO. However, HP affect the performance very much [cite], searching for a proper combination that works optimally for the specific problem can be challenging. Futhermore, in real-world application scenarios, the inference speed is important. There is a tradeoff between the model expressivity and complexity, affecting the performance and inference time. In ocean modeling, the domain is often high dimensional with various variables describing the states interested. In particular, in this work, focus on modeling the ocean dynamics with 5 proganstic variables using FNO. 

We construct the model with Fourier neural operators~(FNOs)~\cite{li2020fourier} and utilize DeepHyper~\cite{balaprakash2018deephyper,deephyper_software} to perform hyperparameter and architecture searches to automate the model selection of FNOs for an idealized baroclinic wind-driven ocean system. The aim is to predict four prognostic variables at a single time step  in the future, given the variable values and a physical parameter at the current time. We also anticipate an improvement in autoregressive rollout performance, which typically requires special treatment during training~\cite{lam2023learning, pathak2022fourcastnet, bi2022panguweather}. This expected enhancement is due to the selected model's ability to provide more accurate forecasts for single-time steps. We  investigate the potential disadvantages of pure MSE loss and the effect of combining MSE and negative ACC with an associated hyperparameter, weighing the importance of the two.
%Our goal is to enhance the model on three fronts: the precision of predictions as measured by loss function values, the autoregressive capability of the model, and the efficiency of inference time. 
Furthermore, we have expanded the hyperparameter optimization (HPO) across a high-performance computing (HPC) environment involving more than 20 nodes, each equipped with four GPUs per node, while integrating effective early stopping criteria to refine the search process, yielding more timely and potent outcomes. The experimental results show that using the optimal hyperparameter configuration from the search results in a model that outperforms the baseline in the single time-stepping forecast of all four prognostic variables and significantly improves the model's autoregressive performance for long-range rollout. We expect this work to pave the way for a more systematic model construction process for ocean/climate modeling using FNOs and their variants. 

The rest of the paper is organized as follows. Section~\ref{sec:related} introduces the work involving deep learning studies for ocean modeling, FNOs, and the general hyperparameter search processes with multiple objectives. Section~\ref{sec:method} formulates the problem into an operator learning setting and describes Bayesian optimization with multipoint acquisitions for parallelization. Sections~\ref{sec:experiment} and \ref{sec:result} define the hyperparameter search space and discuss the search results and the impact of hyperparameters associated with data preprocessing, neural architecture, and training process.  Section~\ref{sec:conclusion} summarizes our conclusions  and briefly mentions future directions.
\section{Related Work}\label{sec:related}
\subsection{Fourier Neural Opearator}
The Fourier neural operator and its variants~\cite{li2020fourier, guibas2021adaptive, li2023physicsinformed, fanaskov2022spectral} have shown outstanding ability to solve partial differential equations and predictive tasks in various scientific domains~\cite{Grady2022Towards, Zhang2023LearningTS}. In particular, FNOs powered climate modeling and enabled accurate mid- and long-range prediction at a fraction of the cost compared with first-principles-based simulations~\cite{pathak2022fourcastnet, bire2023ocean}. The general idea behind FNOs is to composite convolutional kernel integrals  in the Fourier domain to approximate the solution operator. FNOs generally have three major components: lifting layers, Fourier operator blocks, and projection layers. During lifting, the network transforms the input functions~(e.g., initial condition or other parameters) to higher-dimensional spaces. The transformed input goes through a series of FNO blocks where fast Fourier transform~(FFT) transforms the data, followed by convolutional operations in the frequency domain. A preset number of high-frequency modes in the results are filtered out, preserving general features and smoother behaviors associated with lower-frequency modes. An inverse FFT then transforms the data back to the physical domain. Finally, the projection components map the data back to its original dimensions, forming the solution. As an operator learning framework, FNOs learn the mapping between function spaces and can predict beyond the resolution of the training data. Therefore, they are suitable for solving problems involving partial differential equations and various mesh representations of the domain, including weather and ocean modeling.

\subsection{Hyperparameter Optimization with DeepHyper}

DeepHyper~\cite{deephyper_software} is a Python package that provides asynchronous parallel hyperparameter and neural architecture search algorithms.
It provides parallel implementations of the SMAC (sequential model-based algorithm configuration  algorithm~\cite{hutter2011smac}, which is a type of Bayesian optimization algorithm~\cite{jones1998efficient}.
For parallelization, it can leverage different backends with minimal code changes, such as threads, processes, and MPI.
%GAIL - processes seems too vague. I understand there are backend processes but this doesn't sound right
We will now provide an overview of the hyperparameter optimization algorithms we used in this work.

\subsubsection{Centralized Bayesian Optimziation}

Bayesian optimization~(BO), also known as efficient global optimization ~\cite{jones1998efficient}, is a popular approach for hyperparameter optimization frameworks because of its fast convergence properties.
BO is a type of black-box optimization algorithm where the core idea is to iteratively update an internal surrogate model in order to minimize the number of direct queries to the real ``expensive'' black-box function.
The standard BO algorithm is sequential, thus making it difficult to leverage parallel computing resources available on HPC systems.
Within DeepHyper, BO can be parallelized through a centralized or decentralized architecture. In the centralized scheme, a single manager decides the function evaluation locations when workers conduct the evaluations in parallel. In contrast, the decentralized scheme equips workers with their own optimizer and allows all operations in parallel. 
The latter provides maximum scalability~\cite{egele2023asynchronous} when performing large-scale hyperparameter optimization (i.e., > 1,000 parallel function evaluations).
In our case, we use the centralized approach with a $q$UCB ($q$-Upper Confidence Bound) acquisition function~\cite{wilson2017reparameterization} for the multipoint acquisition strategy.
This method provides better scalability than does the usual constant-liar multipoint acquisition strategy~\cite{snoek2012practical}, also known as Kriging Believer~\cite{ginsbourger2010kriging}.
%GAIL - I am a bit confused here. If the centralized approach give maximum scalability, why not use it? You compare the decontralized using qUCB and say you use it for better scalability than the consant-liar (but is that a centralized or decentralized?

In a centralized architecture, the central coordinator determines the next hyperparameters at which the function should be tested and then distributes these tasks to available remote workers.
The surrogate model used is Extremely Randomized Trees~\cite{geurts2006extremely},
 a type of random forest approach~\cite{breiman2001random} that provides better epistemic uncertainty estimates thanks to a random-split strategy when building tree nodes.

\subsubsection{Multiobjective Optimization}

The most common objective in hyperparameter optimization for neural networks is to minimize the validation loss. In our problem setup,  the loss function contains two terms, MSE and negative ACC, and we plan to investigate any potential trade-off between the two.
\romain{maybe cite a paper talking about low-frequency/high-frequency tradeoffs in CFD/ocean modeling?}
%\romain{was (mean-squared-error) MSE or (...) ACC introduced before?}
MSE accounts for the average behavior of the state variable fields, penalizing large errors, whereas ACC focuses on anomalies between forecast and true fields, emphasizing pattern nuances. 
In ocean modeling, we not only care about accurate forecasts of the mean field over time, for example,  the gyre circulation, but also are interested in the precise characterization of the local variations, for example, intergyre exchange from eddies.
We investigate the relationship between the two objectives through the lens of multiobjective optimization.

%%% Explain why and what we expect in the investigation
A trade-off between MSE and ACC would mean that on the optimal frontier, the Pareto front, one objective cannot be improved without degrading the other.
On the contrary, without a trade-off,  both objectives can be improved jointly, thus providing a dominant point on the Pareto front~\cite{kadlec2016multi}, that is, in the objective space.
A simple way to find the Pareto optimal solutions is to combine the objectives and  perform optimization only over the combined single objective, known as scalarization~\cite{ehrgott2005multicriteria}.
 With a trade-off between objectives, however, this process corresponds to optimizing for a fixed trade-off.
DeepHyper can perform multiobjective hyperparameter optimization through randomized scalarization within Bayesian optimization~\cite{egele2023dmobo}.
Randomized scalarization resamples random objective weights for each new suggestion,
thus reducing the multiobjective optimization problem to a single-objective optimization problem where each weight vector represents a fixed trade-off between objectives.
By resampling different weights, the Pareto-Front can be explored.
We leverage this capability to optimize for both validation MSE and ACC.

\subsection{Data-Driven Ocean Modeling}
%The ocean processes are important for mitigating anthropogenic climate change by absorbing substantial carbon dioxide and heat. Simultaneously, oceanic processes that redistribute mass, heat, and salt are instrumental in driving phenomena like hurricanes, extreme precipitation, and droughts. 
Conventional first-principles-based simulations of ocean evolution, while being reliable and accurate, are computationally intensive and can become intractable for tasks such as uncertainty quantification, sensitivity analysis, and optimization. With the advance of deep learning models and the computational efficiency once trained, data-driven methods for modeling ocean dynamics have emerged. Some recent developments address spatial-temporal ocean property forecasts~\cite{zrira2024time} and rainfall pattern segmentation over oceans~\cite{colin2024rain}. However, according to the specific application scenario and datasets, building effective models usually involves careful design of the model architecture and significant trial and error to fine-tune model training to achieve better overall performance. This process introduces a new round of computational overhead. To address this issue, we show a workflow that leverages FNO and hyperparameter optimization to streamline the modeling process of ocean dynamics, which other data-driven models can easily adapt.

\section{Method}\label{sec:method}
\subsection{Problem Formulation}
The dynamics of the idealized baroclinic wind-driven ocean model can be formulated as an initial value problem~(IVP), which starts with the initial state, $\vect{x}_0$, and the model parameter(s), $\kappa$. We aim to solve for the state $\vect{x}(t)$ (i.e., the state prognostic variables) at time $t$. The IVP,
$d\vect{x}(t, \kappa)/dt = f(\vect{x}(t))$, with $\vect{x}(0) = \vect{x}_0$, leads to 
the solution at time $t$,  
    $\vect{x}_t = \vect{x}_0 + \int_{0}^{t} f(\vect{x}(\tau, \kappa))d \tau.$
The solution at the discrete time step $t+1$ can be expressed as 
\begin{equation}\label{eqn:one-step}
    \vect{x}(t + 1) = \vect{x}(t) + \int_{t}^{t+1}f(\vect{x}(\tau, \kappa))d \tau.
\end{equation}
We use a FNO, $\mathcal{N}_{\theta}$, parameterized by learnable weights and biases $\theta$ to approximate the solution operator to (\ref{eqn:one-step}). In particular, we trained the neural network to approximate the following mapping,
\begin{equation}
    \vect{x}_{t+1} = \mathcal{N}_{\theta}(\vect{x}_t, \kappa),
\end{equation}
where $\vect{x}$ is the state variable vector, $t$ represents the current time, and $\kappa$
%\todo{Inconsistent notation: here $\vect{p}$, but in the equations above it is just $p$} 
is the parameter that impacts the trajectories of state variables. With the training dataset $\mathcal{D}$ containing input-output pairs, the learning objective is to minimize the empirical loss function,
\begin{equation}\label{eqn:loss}
\begin{aligned}
    \mathcal{L}({\theta, \mathcal{D}}) &= \alpha \cdot \mathcal{L}_{\text{MSE}} + (1 - \alpha) \cdot \mathcal{L}_{\text{NegACC}},\\
    \mathcal{L}_{\text{MSE}} &= \frac{1}{N}\sum_{i=1}^N \Vert \vect{x}_{t+1}^{(i)} - \mathcal{N}_{\theta}(\vect{x}_t^{(i)}, \kappa^{(i)})\Vert^2_2, \\
    \mathcal{L}_{\text{NegACC}} &= -\frac{\sum_{i=1}^N(\mathcal{N}_{\theta}(\vect{x}_t^{(i)}, \kappa^{(i)}) - \bar{\vect{x}}_{t+1})(\vect{x}^{(i)}_{t+1} - \bar{\vect{x}}_{t+1})}{\sum_{i=1}^N \sqrt{(\mathcal{N}_{\theta}(\vect{x}_t^{(i)}, \kappa^{(i)}) - \bar{\vect{x}}_{t+1})^2(\vect{x}^{(i)}_{t+1} - \bar{\vect{x}}_{t+1})^2}},
\end{aligned}
\end{equation}
where $N$ is the number of data points in the training set. The loss function contains two terms: the standard MSE loss and the negative ACC. We assign a weighting factor $\alpha$, treated as a hyperparameter, to adjust the relative importance between the two terms. We expect the trained neural network to predict the state variables one step forward accurately. 

\subsection{Bayesian Optimization and Multipoint Acquisition}
Bayesian optimization is an efficient global optimization method for black-box functions with the presence of noise. Our HPO problem focuses on finding a set of hyperparameters that maximize the preset objectives. The black-box function, $f(x)$, in this context, is the function that maps hyperparameters to the objectives. Formally, we aim to solve the following,
\begin{equation}
    \max_{p}\{f(p): p = (p_D, p_N, p_T) \in \mathcal{P}\},
\end{equation}
where $p$ is a set of hyperparameters and subscripts $D$, $N$, and $T$ represent hyperparameters in the data preprocessing, neural architecture, and training process, respectively. The hyperparameter search space $\mathcal{P}$ is predefined. The evaluation of $f(p)$ is computationally expensive because it requires complete training of neural networks. Therefore, a surrogate model is needed to approximate these expensive functions, allowing for a more efficient exploration of the hyperparameter space. 
We used the default surrogate model in DeepHyper, random forest~\cite{breiman2001random}, to perform BO. Random forest regressors have scaling advantages over more commonly used Gaussian processes~\cite{egele2023asynchronous}. 

In BO, the acquisition function describes the quality of the input and determines the next evaluation points. We used the upper confidence bound~(UCB), defined as the follows,
\begin{equation}
    UCB(p) = \mu(p) + c \cdot \sigma(p),
\end{equation}
where $\mu(p)$ and $\sigma(p)$ are the mean and standard deviation of the surrogate predictions, respectively. The constant, $c$, controls the trade-off between exploration and exploitation. BO iteratively evaluates $f(p)$ and determines the next evaluation point(s) $p^\prime$ based on the value of $UCB(p)$. 

To perform efficient searches and leverage the HPC environment, we enabled multipoint acquisition with UCB to parallelize BO. This poses a multipoint optimization problem. In the centralized scheme, a manager performs BO and selects hyperparameter configurations, and workers evaluate the configurations and return the results back to the manager. Since BO typically updates the surrogate model and selects points sequentially, we used the $q$UCB strategy~\cite{wilson2017reparameterization} to ensure effective evaluation point selection in parallel, where various values of $c$ are drawn from the exponential distribution for the UCB acquisition function and different next points are chosen according to these values to achieve an optimal balance between exploration and exploitation.

\section{Experiment}\label{sec:experiment}
This section describes the experiment setup, including the data generation, hyperparameter search space, and optimization execution.

\subsection{Dataset}
We adopted and simplified the data generated from \cite{sun2023surrogate}.
The dataset used in the experiment consisted of 100 simulations of salinity, temperature, zonal velocity, and meridional velocity at the ocean surface of an idealized baroclinic wind-driven ocean model. They were obtained by running the Simulating Ocean Mesoscale Activity~(SOMA) test case~\cite{wolfram2015diagnosing} within a circular basin of a 150 km wide and 100 m deep shelf. An ensemble of simulations was developed by varying the bolus diffusivity ($\kappa_{GM}$) in the Gent--McWilliams parameterization. The diffusivity was uniformly sampled from $[200, 2000]$. Each variable of interest has two dimensions in space, spanning 100 grid points per direction. Therefore, along with $\kappa_{GM}$, the data obtained from each simulation~(associated with a different $\kappa_{GM}$) has the shape of $(30, 100, 100, 5)$, where 30 indicates 30 time steps~(days) and 5 includes the four state variables and $\kappa_{GM}$. To train models that take the state variables and parameters at the current time and predict the same set of state variables at the next step, we split 30 time steps into 29 input-output pairs per simulation, resulting in 2,900 input-output instances in total. We further split the data based on independent simulations into training, validation, and testing sets with a ratio of 0.6, 0.2, and 0.2, respectively. The search for the optimal set of hyperparameters was determined by the model performance on the validation set. The final evaluations were done using the testing set.

\subsection{Hyperparameter Search Space}
The hyperparameters involved in this work mainly belong to three categories: data preprocessing, neural architecture related, and training process related. The neural-architecture-related hyperparameters determine the hypothesis~(model) space, bounding the expressivity, and generalizability. The training-process-related hyperparameters affect the learning, of which the goal is to find the optimal model by varying its trainable weights to minimize the loss function over the training and validation data. We focus on searching for the optimal set of hyperparameters, listed in Table~\ref{tab:hp}, that contribute to the best-performing FNO for ocean modeling. 
% \begin{itemize}
%     \item Number of FNO layers: number of spectral convolutional layers.
%     \item FNO layer size: feature size in the spectral convolutions.
%     \item Number of FNO modes: number of Fourier modes to keep after filtering.
%     \item Padding: the padding size for the input domain.
%     \item Padding type: type of padding to perform.
%     \item Activation function: choice can be from all torch-supported activation functions.
%     \item if using the coordinate feature as an additional feature map. 
%     \item Number of decoder layers. (Decoder is the projection operation in FNO).
%     \item Decoder layer sizes.
%     \item Decoder layer activation
% \end{itemize}
% Hyparameters in training
% \begin{itemize}
%     \item Number of epochs.
%     \item Optimizer
%     \item Learning rate.
%     \item Regularization.
%     \item Learning rate scheduler.
%     \item Batch size.
%     \item 
% \end{itemize}
% The objective is to minimize the validation loss. We can consider secondary objectives, e.g., training time, number of trainable parameters, and inference time.
\begin{table}[]
    \centering
    \caption{List of hyperparameters and their ranges. (a) shows the hyperparameters related to data preprocessing. (b) shows the hyperparameters determining the neural architecture. (c) describes the ones in the training process.}

    \begin{subtable}{\linewidth}
    \caption{}
    \resizebox{\linewidth}{!}{
    \begin{tabular}{p{3cm}p{1cm}p{5cm}p{5cm}}
    \toprule
        Variable Names.      & Type & Range/Choice                                   & Explanation\\
    \midrule
        \hp{padding}         & \hp{bool} & \hp{[True, False]}                                  & If zero-pad the data. \\
        \hp{padding\_type}   & \hp{str}  & \hp{[`constant', `reflect', `replicate', `circular']}                      & Types of padding.\\
        
        \hp{coord\_feat}     & \hp{bool} & \hp{[True, False]}                                  & If use domain coordinates as additional features.\\
    \bottomrule
    \end{tabular}
    }
    \end{subtable}

    \begin{subtable}{\linewidth}
    \caption{}
    \resizebox{\linewidth}{!}{
    \begin{tabular}{p{3cm}p{1cm}p{5cm}p{5cm}}
    \toprule
        Variable Names                                        & Type & Range/Choice                                   & Explanation\\
    \midrule
         \hp{lift\_act}            & \hp{str}  & \hp{[`relu', `leaky\_relu', `prelu', `relu6', `elu', `selu', `silu', `gelu', `sigmoid', `logsigmoid', `softplus', `softshrink', `softsign', `tanh', `tanhshrink', `threshold', `hardtanh', `identity', `squareplus']}                   & Activation function for lifting layers. The choices include common activation functions implemented in \texttt{PyTorch}.\\
         
        \hp{num\_FNO}              & \hp{int}  & [2, 16]                                        & The number of FNO blocks.\\
        \hp{num\_latent\_feat}                          & int  & [2, 64]                                        & The number of latent features in FNO blocks. This is equivalent to the number of channels in an image representation.\\
        
        \hp{num\_modes}           & \hp{int}  & [2, 32]                                        & The number of Fourier modes to keep.\\
        \hp{num\_proj\_layers}    & \hp{int}  & [2, 16]                                        & The number of projection layers. \\
        \hp{proj\_size}           & \hp{int}  & [2, 16]                                        & Projection layer size. \\
        \hp{proj\_act}                    & \hp{str}  & \hp{[`relu', `leaky\_relu', `prelu', `relu6', `elu', `selu', `silu', `gelu', `sigmoid', `logsigmoid', `softplus', `softshrink', `softsign', `tanh', `tanhshrink', `threshold', `hardtanh', `identity', `squareplus']}                   & Activation function for projection layers. The choices include common activation functions implemented in \texttt{PyTorch}.\\
        
    \bottomrule
    \end{tabular}
    }
    \end{subtable}

    \begin{subtable}{\linewidth}
    \caption{}
        \resizebox{\linewidth}{!}{
    \begin{tabular}{p{3cm}p{1cm}p{5cm}p{5cm}}
    \toprule
        Variable Names                                                                                                                                                                                                                  & Type & Range/Choice                                   & Explanation\\
    \midrule
        \hp{alpha}          & \hp{float} & (0, 1) & Weight associated with MSE and negative ACC in the loss function.\\
        \hp{optimizer}      & \hp{str}  & \hp{[`Adadelta', `Adagrad', `Adam', `AdamW', `RMSprop', `SGD']} & Types of optimizers.\\
        \hp{lr}             & \hp{float} & (1e-6, 1e-2)                                  & Learning rate\\
        \hp{weight\_decay}  & \hp{float}  & (0, 0.1)                  & The weighting factor of the $L_2$ regularization.\\
        \hp{batch\_size}    & \hp{int}  & (2, 64)                     & The batch size of training data during training.\\
    \bottomrule
    \end{tabular}}
    \end{subtable}
    \label{tab:hp}
\end{table}

% \begin{table}[]
%     \centering
%     \caption{Hyperparameter list related to the model training process.}
%     \resizebox{\linewidth}{!}{
%     \begin{tabular}{lccc}
%     \toprule 
%         Variable Names &  Type & Range/Choice & Explanation\\
%     \midrule
%         epochs & bool & [True, False] & The number of epochs for training \\
%         Optimizer & str & [Adadelta, Adagrad, Adam, AdamW, RMSprop, SGD] & Types of optimizers.\\
%         lr & bool & [True, False] & Learning rate\\
%         regularizer & str & [relu, silu, tanh, identity] & Types of regularization.\\
%         lr sch & int & (2, 16) & The learning rate scheduler to use\\
%         batch size & int & (2, 64) & The batch size for training and validation data.\\
%     \bottomrule
%     \end{tabular}}
%     \label{tab:my_label}
% \end{table}
Since the convolution operations in the Fourier domain are the critical components of FNOs, the usage of padding and coordinate features is essential to the learning results. The first set of hyperparameters is whether we use padding for the input data and the padding type. 
Paddings affect the output of the convolutions at the domain boundaries~\cite{alrasheedi2023padding} and potentially mitigate the edge effect introduced by the subsequent FFT. Also, we 
consider incorporating the coordinate features on the input data level, allowing the model to learn the explicit solution dependency on the spatial position.

In the lifting block of FNOs, which maps the input data to higher dimensions elementwise~(preserving the spatial dimension), we aim to optimize the number of layers and the activation functions. 
The next component in FNOs is the operator learning block, consisting of the forward FFT, convolution operation in the frequency domain, filtering out of high-frequency modes, followed by nonlinear activations, and then an inverse FFT. Here, we search for the optimal number of operator learning blocks, the number of Fourier (low-frequency)~modes to keep, and the choice of the activation function. The last component in the typical FNOs is the projection block, transforming the lifted and convolved data back to the original data dimension. Similar to the lifting block, we are interested in finding the optimal number of projection layers, the number of neurons per layer, and the choice of the activation function.
During training, the choice of the optimizer, batch size, magnitude of the learning rate, weights associated with the loss terms, and regularization could highly impact the quality of the trained model. Therefore, we aim to find an optimal combination of these values that leads to the most accurate model.
\subsection{Objectives}
We set up the hyperparameter optimization problem to optimize two objectives. The  first objective is to minimize the validation MSE, defined as the first term in (\ref{eqn:loss}). MSE is calculated pixelwise, aiming to reach the average minimum value. The second objective is to maximize the validation anomaly correlation coefficient~(ACC), defined as the second term in (\ref{eqn:loss}), which is a commonly used metric in climate studies. Given that adding negative ACC could potentially avoid the drawbacks of MSE, we treat them as two separate objectives with equal importance and investigate the relationship between ACC and MSE. In DeepHyper, all the searches are focused on maximizing objectives. Therefore, we implemented the objectives as negative MSE and positive ACC.
% While training using mean squared error~(MSE) as the loss function, previous experiments show that if keeping MSE as the objective for model selection, the HPO process tends to limit the number of Fourier modes to keep to a lower number, eliminating higher frequency features. Therefore, we use ACC, accounting for the anomalies~(profile differences from the mean field), as one of the objectives of selecting models. The second objective is to minimize the inference time. Inference time is counted as the average time spent for a trained model from taking a single input to producing the output. A lower inference time is usually associated with a less complex neural network structure, so we expect to see a trade-off between the network performance and inference time. The last objective is to maximize the rollout of ACC. In the neural network setup, we train it for one-time-stepping forecasting, taking the state variable and parameter at the current time to predict the same set of state variables at the next time step. When used autoregressively, such models tend to cause fast error accumulation. This can be addressed by training the network for multiple steps, which will increase the computational overhead. In this work, we aim to mitigate this issue via HPO by adding the rollout ACC as one of the optimization objectives.

\subsection{Implementation}
We constructed the FNO model using the \texttt{Nvidia Modulus} package and set up the hyperparameter search space, the black-box function for BO, and the evaluator in \texttt{DeepHyper}. The searches used 20 computing nodes, with four Nvidia A100 GPUS per node, on \texttt{Polaris} cluster at the Argonne Leadership Computing Facility.

Training FNO with high-dimensional data can be computationally expensive. Therefore, we adopted two stoppers to accelerate the searches and the parallelization. A stopper terminates the training of a neural network based on certain criteria, improving search efficiency. The first stopper used a constant predictor, a constant number that minimizes the MSE term in the loss function (\ref{eqn:loss}). In this case, the constant predictor becomes the mean value of the target. The second stopper keeps track of the running time for each training epoch. In our optimized training regimen for each hyperparameter configuration, we implemented an early stopping mechanism for efficiency. Specifically, we terminated the training of any models whose configurations yielded results inferior to that of the constant predictor within the initial 10 epochs or taking over 100 seconds to complete a single epoch during training. If neither stopper was triggered, the training was terminated at epoch 30, returning the objective values. This approach strategically reduces unnecessary computational expenditure on unpromising model configurations.

With the search results, we conducted a full training~(100 epochs) using the set of hyperparameters that resulted in the best objectives. We used the default hyperparameter configuration preset in \texttt{Modulus} as a baseline to compare the testing performance with the fully trained model.

\section{Results and Discussion}\label{sec:result}

To demonstrate the positive impact of using negative ACC as the additional loss term, we trained models using the default hyperparameters provided by Modulus for 100 epochs using loss functions that are pure MSE and the equally weighted ones of MSE and negative ACC. While the search objectives were validation MSE and ACC, we use Relative Squared Error~(RSE), $RSE = \sum(\vect{x}_{t+1} - \mathcal{N}_{\theta}(\vect{x}_t, \kappa))^2/\sum(\vect{x}_{t+1} - \bar{\vect{x}}_{t+1})^2$, and 1 - ACC in their log scale to highlight performance differences. We also use these metrics to evaluate search results in the later sections. 

Table~\ref{tab:baselineCompare} shows the $\log(RSE)$ and $\log(1-ACC)$ values from the two models. Overall, the model trained with the composite loss function achieved lower values for both metrics. Notably, training the network with the composite loss improved MSE values over the pure MSE loss in salinity, meridional velocity, and zonal velocity. 
%\textit{The performance of forecasting layer thickness worsened in RSE but drastically improved in 1-ACC. This could be caused by the fact that in our dataset, layer thickness values are extremely uniformly distributed throughout the domain compared to other variables and have very minimal variance.}  Figure~\ref{fig:baseline_temp} shows the predictions from both models on temperature in the testing set. Both models achieve accurate forecasts where the model trained with the composite loss presents slightly smaller errors, especially near the domain boundaries. 
These results validate adding negative ACC as part of the loss function. The model training for the subsequent hyperparameter optimization adopts this composite loss.
%\todo{some comments on layer thickness}

% The baseline on testing set have average r2 and acc
% [-19.60108308   0.99371335   0.99979882   0.98899044   0.99435678]
% [0.01713433 0.99686127 0.99990351 0.99448192 0.99717445]
% [1.55205374e-04 2.53743851e-06 6.03262873e-03 5.23747266e-05
%  7.76394641e-05]

% The MSE + ACC baseline testing set average r2 and acc
% [-1.29544283e+03  9.99281220e-01  9.99896301e-01  9.98801738e-01
%   9.99029297e-01]
% [0.99938434 0.99971924 0.99994825 0.99944242 0.99955199]
% [9.76719979e-03 2.90116407e-07 3.10952455e-03 5.70037953e-06
%  1.33549520e-05]

% remap it back to the original scale 
% MSe only 
% [0.01713433 0.99686127 0.99990351 0.99448192 0.99717445]
% [1.55205374e-04 2.53743851e-06 6.03262873e-03 5.23747266e-05
%  7.76394641e-05]
% log rse [ 1.31389005 -2.20158099 -3.69641321 -1.95823008 -2.24847331]
% log 1-ACC [-0.00750584 -2.50324541 -4.01553252 -2.25821187 -2.54889666]

% ACC + MSe
% [0.99954281 0.99841561 0.99956578 0.99855982 0.99844529]
% [6.56299964e-02 1.28201235e-06 2.60485731e-02 4.07971263e-05
%  6.84955166e-05]

% log rse [ 3.94008568 -2.49808431 -3.06113587 -2.06672228 -2.30289369]
% log 1-ACC [-3.33990332 -2.80013793 -3.36229269 -2.84158398 -2.80835118]

\begin{table}[]
    \centering
    \caption{Model performance on the testing set using the baseline models trained with MSE loss and composite loss.}
    \resizebox{\textwidth}{!}{
    \begin{tabular}{lcccc}
        \toprule
                             & \multicolumn{2}{c}{$\log(RSE) \downarrow$} & \multicolumn{2}{c}{$\log(1 - ACC)\downarrow $}\\
        \midrule
                             & MSE Loss                         & MSE + NegACC Loss                  & MSE Loss & MSE + NegACC Loss\\
        \midrule
       % Layer Thickness      & $\mathbf{1.314}$              & $3.940 $             & $-0.008$           & $\mathbf{-3.339}$\\
        Salinity             & $-2.202$                       & $\mathbf{-2.498}$   & $-2.503$           & $\mathbf{-2.800}$\\
        Temperature          & $\mathbf{-3.696}$              & $-3.061$            & $\mathbf{-4.015}$  & $-3.363$\\
        Meridional V.        & $-1.958$                       & $\mathbf{-2.067}$   & $-2.258$           & $\mathbf{-2.842}$\\
        Zonal V.             & $-2.248$                       & $\mathbf{-2.303}$   & $-2.549$           & $\mathbf{-2.808}$\\
        \bottomrule
    \end{tabular}
    }
    \label{tab:baselineCompare}
\end{table}

\begin{figure}
    \centering
    \begin{subfigure}{\linewidth}
        \caption{}\label{fig:data-pc}
        \vspace{.5em}
        \includegraphics[height=3.5cm]{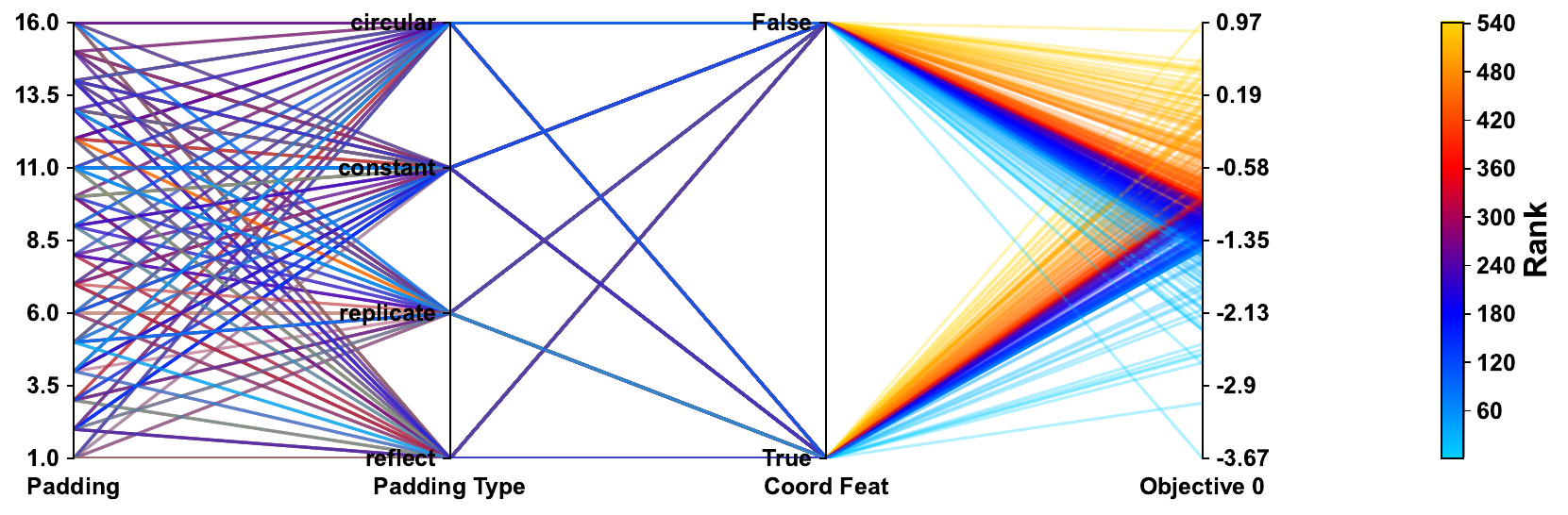}        
    \end{subfigure}
    \begin{subfigure}{\linewidth}
        \caption{}\label{fig:tr-pc}
        \vspace{.5em}
        \includegraphics[height=3.5cm]{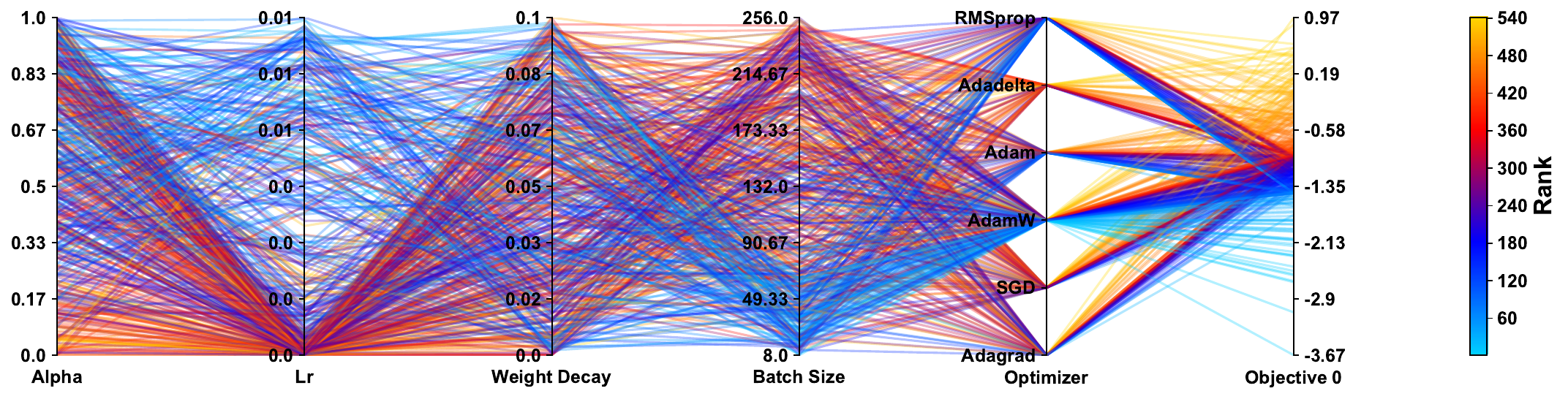}        
    \end{subfigure}
    \begin{subfigure}{\linewidth}
        \caption{}\label{nas-pc}
        \vspace{.5em}
        \includegraphics[height=3.5cm]{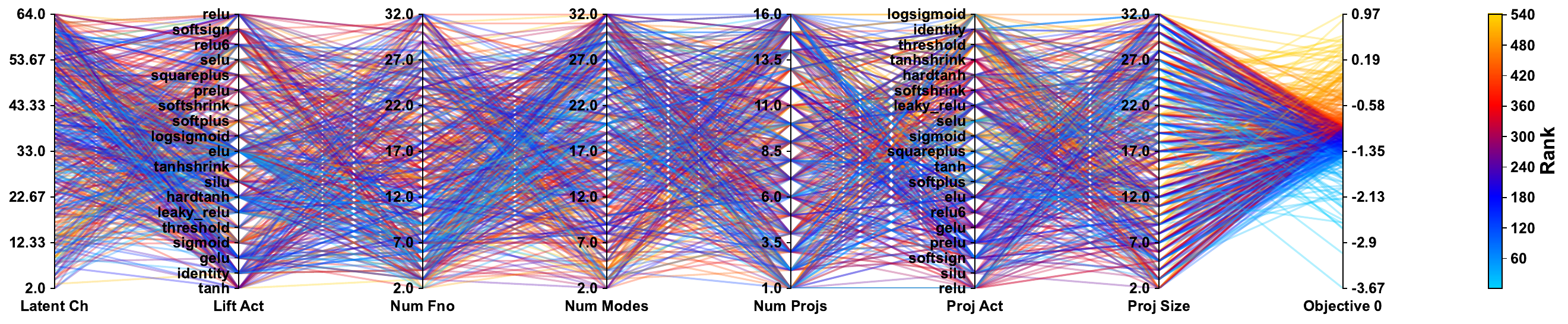}        
    \end{subfigure}
    \caption{Parallel coordinate plots of the hyperparameters in the search space with respect to the validation MSE~(in log scale). The space is divided into three categories. (a) shows the data-related hyperparameters, (b) shows training-related hyperparameters, and (c) contains neural architecture-related hyperparameters.}
    \label{fig:pc_plots}
\end{figure}

\subsection{First Objective: Validation Mean Square Error}
Varying separate hyperparameters results in differing responses in the validation MSE. Figure \ref{fig:pc_plots} shows the parallel coordinates plots of the hyperparameters in the search space, divided into three categories. The curves are ranked based on the objective values in the log scale, where the lighter blue represents favorable configurations. Figure \ref{fig:data-pc} suggests that, among the data-related hyperparameters, in this case, the padding (its type and coordinate features, if used) has an unclear effect on the final model performance in validation MSE. The reason could be that the region of interest is a circular basin represented in the regular grids, with the area outside the basin being zeros, and using padding or different types of padding only expands the area, not affecting learning. Figure \ref{fig:tr-pc} shows the effect of various training strategies. Among the hyperparameters for training, the loss term weighing factor $\alpha$ near the two ends of its range results in suboptimal performance. A smaller batch size corresponds to a lower validation MSE, while the magnitude of the learning rate seems to have negatively affected the validation MSE. We suspect this was due to the limited number of epochs in the search, where models update more slowly with lower learning rates and wind up with worse performance at the end of training. The weight decay does not show an evident correlation with the model performance. A smaller batch size is considered useful for improving generalization error and accelerating convergence~\cite{bengio2012practical}, coinciding with the search results.  The configurations with lower validation MSE are present  mostly with the AdamW optimizers, while Adadelta is associated with mostly lower-ranked configurations. The Adam optimizer~\cite{kingma2017adam}, the default optimizer in our cases, has led to mixed model performances. 
Regarding the neural architecture-related hyperparameters, we can observe evident impact from the choice of the number of latent channels, number of FNO blocks, number of Fourier modes, number of projection layers, and projection activation functions. More specifically, latent channels being greater than two and less than ten, fewer FNO blocks, a higher number of Fourier modes, and a mid- to low-level number of projection layers correspond to a lower validation MSE. Among the projection activations, the commonly used rectified linear unit~(ReLU) led to relatively suboptimal performance. However, some ReLU variants---leaky ReLU and PReLU---positively influenced the minimization of the validation MSE.

\begin{figure}
    \centering
    \begin{subfigure}{\linewidth}
        \caption{}
         \vspace{.5em}
         \includegraphics[height=3.5cm]{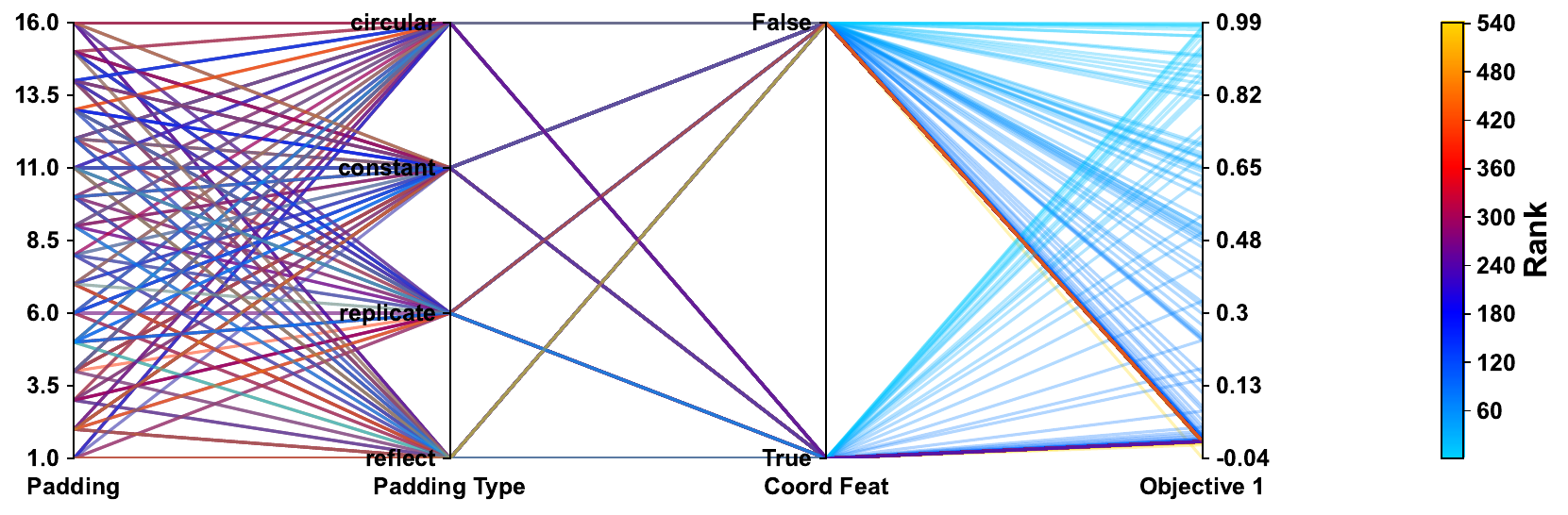}
         \label{fig:acc_data}
    \end{subfigure}

    \begin{subfigure}{\linewidth}
        \caption{}
        \vspace{.5em}
         \includegraphics[height=3.5cm]{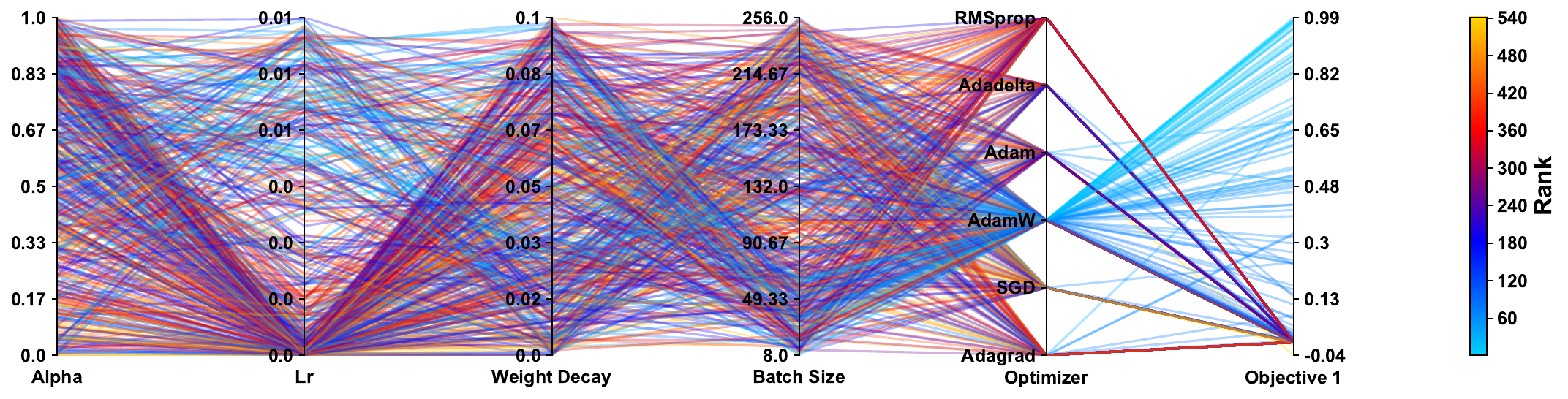}
         \label{fig:acc_train}
    \end{subfigure}
    
    \begin{subfigure}{\linewidth}
        \caption{}
        \vspace{.5em}
         \includegraphics[height=3.5cm]{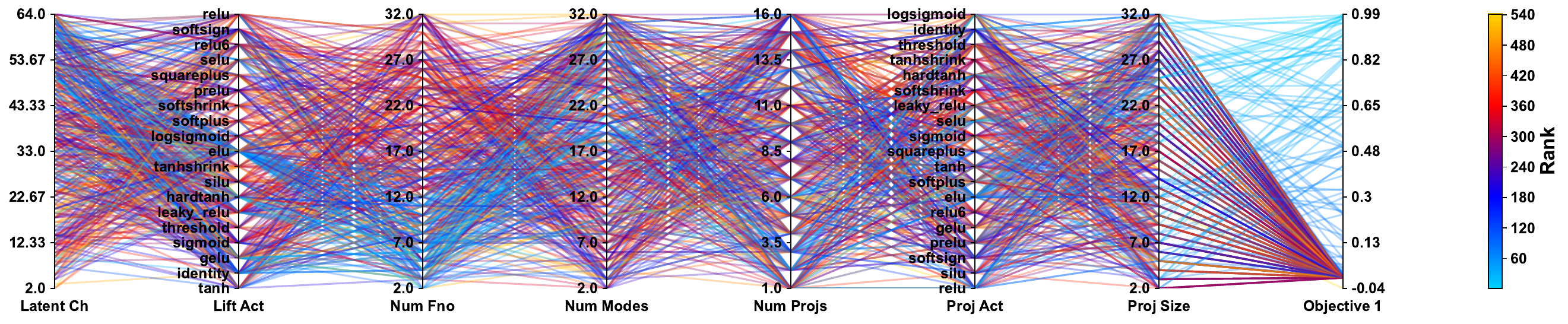}
         \label{fig:acc_nas}
    \end{subfigure}
    \caption{Parallel coordinate plots of the hyperparameters in the search space with respect to the validation ACC. The space is divided into three categories. (a) shows the data-related hyperparameters, (b) shows training-related hyperparameters, and (c) contains neural architecture-related hyperparameters.}
    \label{fig:pc_obj1}
\end{figure}

\subsection{Second objective: Validation Anomaly Coefficient Correlation}
The second search objective was the validation ACC, which measures how much the forecasted deviation from the mean field is correlated with the true deviation. A higher ACC is preferred; thus, the search aimed to maximize the ACC. Figure~\ref{fig:pc_obj1} shows the parallel coordinates plots of the impact of hyperparameters on the validation ACC, where the lines are ranked with decreasing ACC values. Different hyperparameters have different impact. Figure~\ref{fig:acc_data} implies that the data preprocessing approaches do not have an obvious impact on the validation ACC, which coincides with the case for validation MSE. In Figure~\ref{fig:acc_train} the loss weighing factor $\alpha$ does not present an obvious impact on the validation ACC. Similar to the impact on the validation MSE, a greater learning rate positively affects the ACC. The influence of the weight decay magnitude is  clearer compared with it on the validation MSE, where a lower weight decay leads to higher ACC.
Conversely, the impact of the batch size is not as evident as it is for the validation MSE, but a smaller value continues to have a positive impact on the performance. Regarding the choice of optimizers, the AdamW optimizer leads to the most high-ranking configurations. Regarding the neural architecture-related hyperparameters, shown in Figure~\ref{fig:acc_nas}, a higher number of latent channels is associated with higher ACC, a trend that is less evident in the validation MSE. Regarding the activation functions in the lifting layer, elu seems to be most helpful, but the advantage is not prominent. Similar to the impact on the validation MSE, fewer  FNO blocks lead to better ACC. Within each FNO block, the number of Fourier modes to keep does not suggest a straightforward pattern, where having  around 20 such nodes results in many highly ranked configurations. The rest of the hyperparameters determining the neural architecture do not present a strong correlation with the validation ACC. 

% \begin{figure}[H]
%     \centering
%     % \begin{adjustwidth}{-\extralength}{0cm}
%     \begin{subfigure}{\linewidth}
%         \caption{}\label{fig:data-pc}
%         \vspace{.5em}
%         \includegraphics[height=4cm]{figs/PC_data_objective-2.pdf}        
%     \end{subfigure}
%     %
%     \begin{subfigure}{\linewidth}
%         \caption{}\label{fig:tr-pc}
%         \vspace{.5em}
%         \includegraphics[height=4cm]{figs/PC_tr_objective-2.pdf}        
%     \end{subfigure}
%     %
%     \begin{subfigure}{\linewidth}
%         \caption{}\label{nas-pc}
%         \vspace{.5em}
%         \includegraphics[height=4cm]{figs/PC_NAS_objective-2.pdf}        
%     \end{subfigure}
%     \caption{Parallel coordinate plots of the hyperparameters in the search space. The space is divided in three categories. (a) shows the data-related hyperparameters, (b) shows training-related hyperparameters, and (c) contains neural architecture-related hyperparameters.}
%     \label{fig:pc_plots}
%     % \end{adjustwidth}
% \end{figure}
\subsection{Relationship between MSE and ACC}
\begin{figure}
    \centering
    \includegraphics[width=\linewidth]{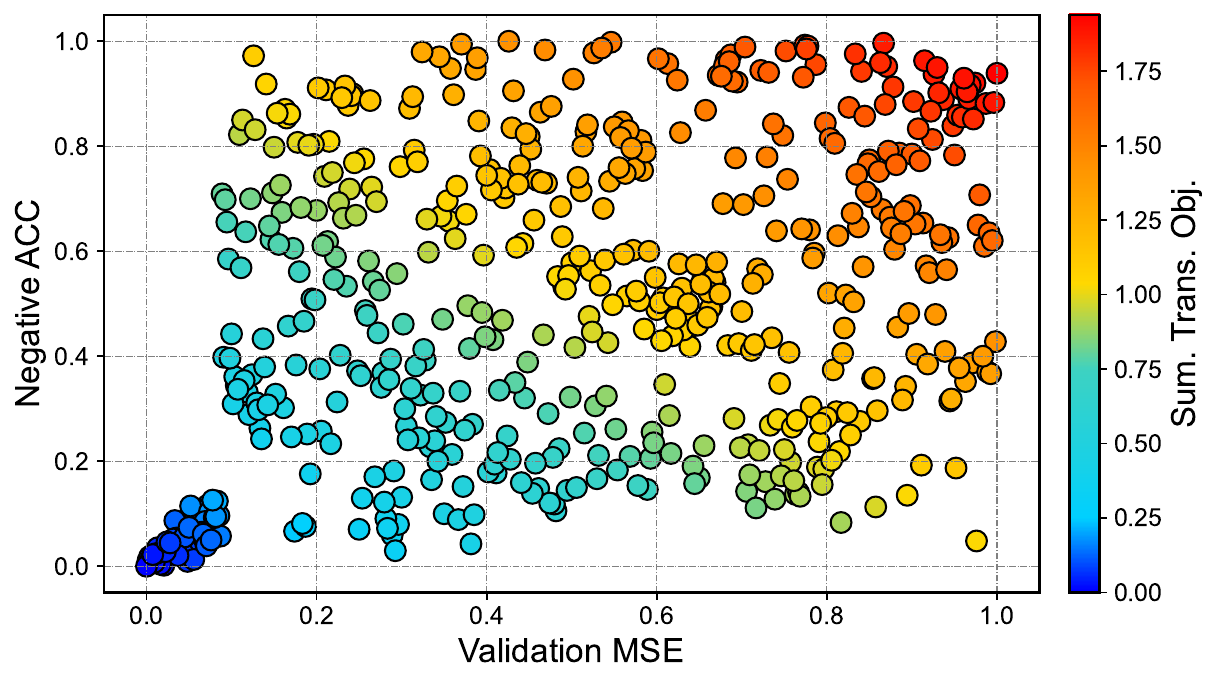}
    \caption{Scatter plot of the quantile-transformed MSE and negative ACC among the search results. The points are color-coded based on the summation of the two objectives, where a lower value indicates better performance. }
    \label{fig:scatter}
\end{figure}

In the search configuration, we sought to optimize for both lower validation MSE and higher ACC. Given that the composite loss improves learning compared with MSE-only loss,  we want to determine whether the two loss terms have any trade-offs. Figure~\ref{fig:scatter} shows the scatter plot of the MSE and negative ACC values among the search results. The values are quantile transformed for easy comparison. The plot suggests no trade-off between the MSE and negative ACC. The optimal configuration in the search results led to the lowest MSE and negative ACC. This aligns with our observation, where negative ACC helped lower MSE in the baseline configuration.  

While the MSE penalizes large errors and focuses on minimizing the average squared error between the forecast and ground truth, ACC pays more attention to the deviation from the mean field, encouraging the model to capture the patterns more accurately. From our experiment, ACC improved overall learning and helped reduce MSE, which might stem from smoothing the optimization landscape or introducing a beneficial gradient that led to easier and faster convergence.  Therefore, the simple addition of negative ACC might benefit ocean modeling or climate modeling in general using FNOs. We plan to investigate the phenomenon and test this hypothesis in future work.

\subsection{Train with Optimal Configuration}
\begin{figure}
    \centering
    \begin{subfigure}{\textwidth}
        \caption{}
        \vspace{.5em}
        \includegraphics[width=\linewidth]{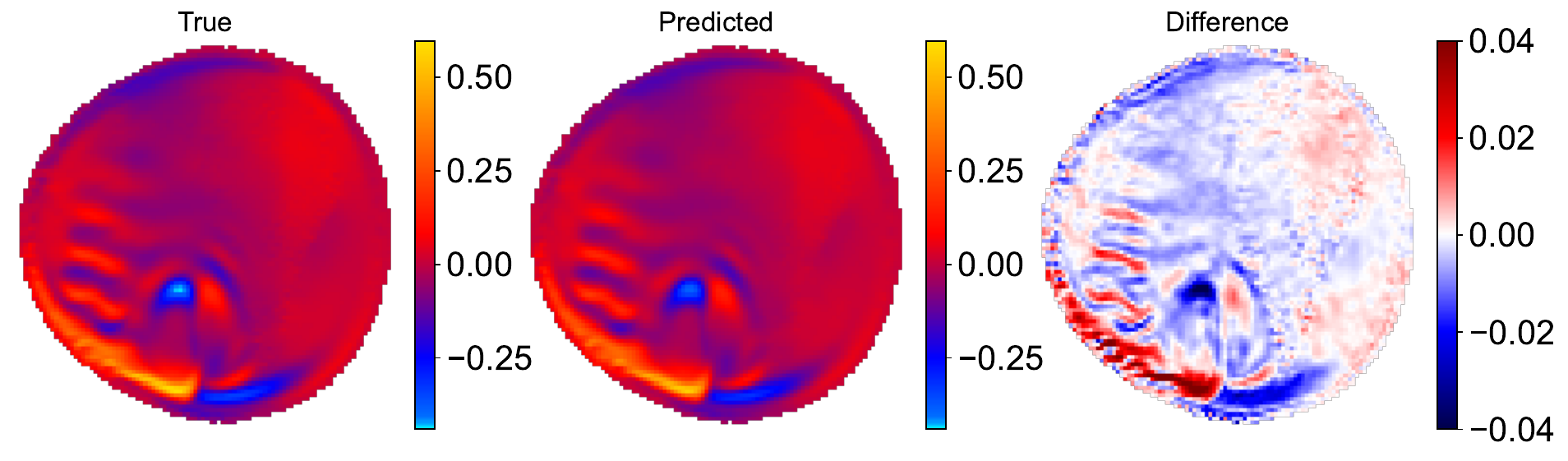}
    \end{subfigure}
    
    \begin{subfigure}{\textwidth}
        \caption{}
        \vspace{.5em}
        \includegraphics[width=\linewidth]{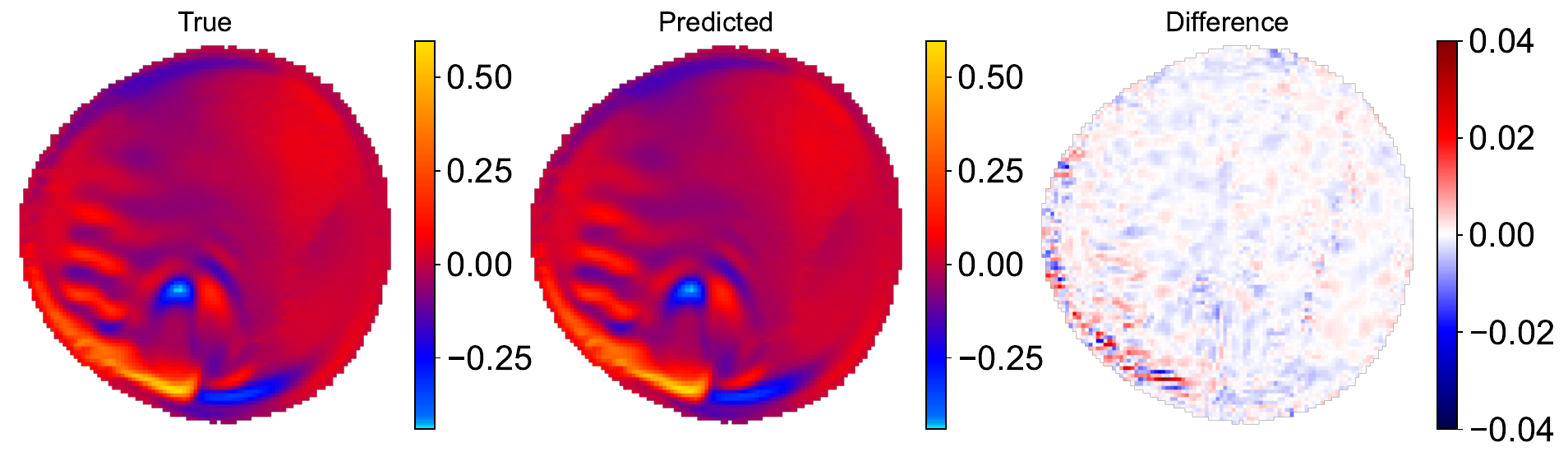}
    \end{subfigure}
    \caption{Meridional velocity profiles of model predictions using the testing set. Both models, with different hyperparameters, were trained with 100 epochs using the composite loss function. (a) shows the model performance with the baseline hyperparameter configuration; (b) shows the model with the best configuration from the search results.}
    \label{fig:temperature}
\end{figure}
With the search results, we selected the hyperparameter configuration that produced the best performance in both validation MSE and ACC and a model for 100 epochs to report its testing performance and autoregressive rollout scores. Table 3 shows the performance among all the variables of interest using the model trained with the baseline configuration and optimal configuration from the search. The optimal configuration outperforms the baseline for all four variables regarding the ACC score, while the baseline has a slight advantage in terms of the MSE for meridional velocity profiles.

\begin{table}
    \centering
    \caption{MSE and ACC scores of the baseline model  with the optimal configuration from the search results.}
    \begin{tabular}{lcccc}
    \toprule
                           & \multicolumn{2}{c}{Baseline} & \multicolumn{2}{c}{Optimal}\\
    \midrule
                           & $\log(RSE) \downarrow$  & $\log(1 - ACC) \downarrow$ & $\log(RSE) \downarrow$  & $\log(1 - ACC) \downarrow$  \\
    \midrule
    % Layer Thickness        &  $3.940$   &  $-3.340$   & $3.113$  & $-3.211$      \\
    Salinity               &  $-2.498$  &  $-2.800$   & $-3.143$ & $-3.552$    \\ 
    Temperature            &  $-3.061$  &  $-3.362$   & $-3.984$ & $-4.286$         \\
    Meridional V.          &  $-2.067$  &  $-2.842$   & $-2.921$ & $-3.254$      \\ 
    Zonal V.               &  $-2.303$  &  $-2.808$   & $-3.013$ & $-3.349$   \\ 
    \bottomrule
    \end{tabular}
\end{table}

\begin{figure}
    \centering
    \begin{subfigure}{.48\textwidth}
        \centering
        \caption{}
        \vspace{.5em}
        \includegraphics[width=\linewidth]{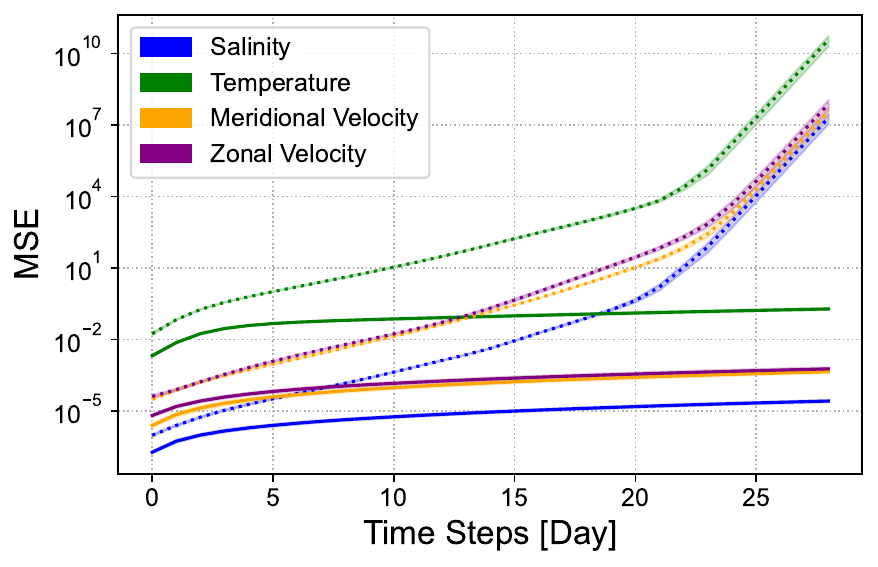}
    \end{subfigure}
    \begin{subfigure}{.48\textwidth}
        \centering
        \caption{}
        \vspace{.5em}
        \includegraphics[width=\linewidth]{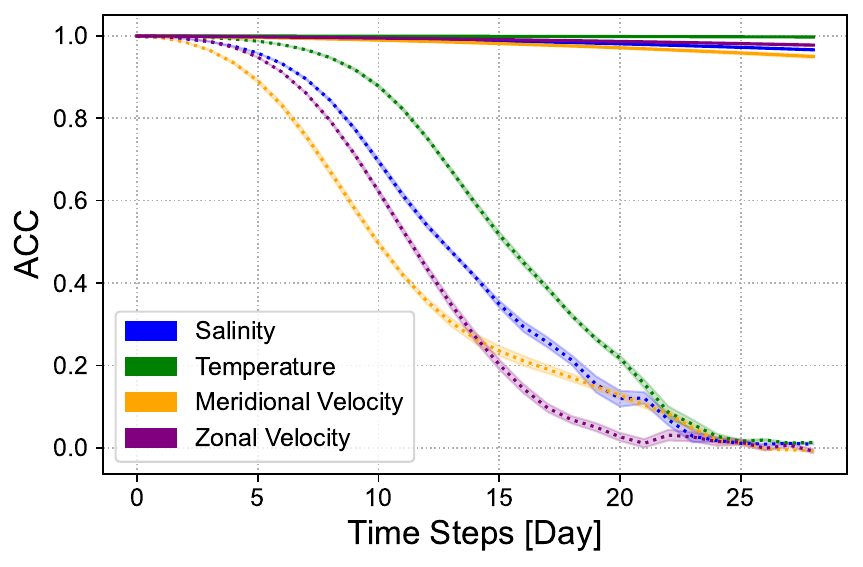}
    \end{subfigure}

    \begin{subfigure}{\textwidth}
        \centering
        \caption{}
        \vspace{.5em}
        \includegraphics[width=\linewidth]{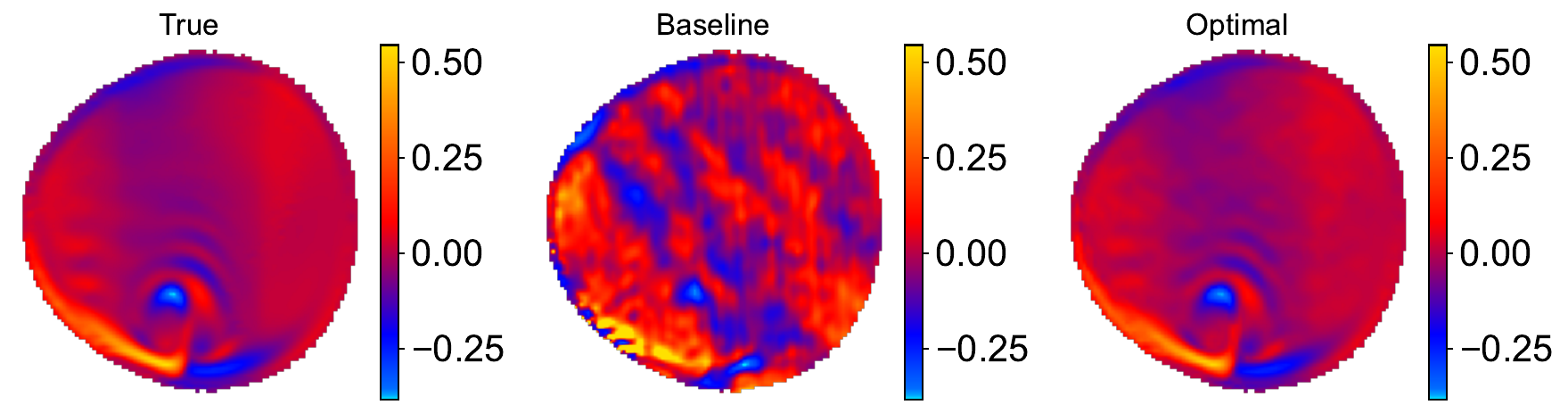}
        \label{fig:meri_rollout}
    \end{subfigure}
    \caption{Rollout performance comparison between the models with baseline hyperparameters and searched optimal configurations. (a) shows the rollout MSE values where the error accumulation is much slower for the model with the optimal configuration; (b) shows the rollout ACC scores where the model with default configuration degrades quickly as the rollout horizon increases; (c) shows the model rollout forecasts at Day 10 for meridional velocity profiles.}
    \label{fig:rollout}
\end{figure}

The slight improvement is amplified by reapplying 
%GAIL - not sure about reapplicating -- reusing? reusing?
the trained models for a longer horizon rollout. Figure~\ref{fig:rollout} shows the model autoregressive rollout performance among the trajectories in the testing dataset. We generated the rollouts by providing the ground truth state variable values at the beginning and used the model to produce the forecast for the next time step. Then, we fed the forecast to the same model as input and obtained the prediction for the next. This process was repeated 29 times, generating forecasting trajectories for an entire month. Both baseline and optimal models start around the same performance. However, the baseline model quickly degrades. Regarding the model with the baseline configurations, the MSE growth rates are similar among the four variables, while meridional velocity shows the earlier ACC degradation and a higher decreasing rate. In comparison, the model using the optimal configurations displays very low error accumulation in MSE and a minimal decrease in the ACC score.  Figure~\ref{fig:meri_rollout} shows the visualization of the rollout meridional velocity profiles on Day 10 from the models with baseline and optimal configurations. The optimal configuration greatly has improved the model's autoregressive performance, allowing more accurate longer horizon forecasts. 

% Best config scores
% [0.99938434 0.99971924 0.99994825 0.99944242 0.99955199]
% [9.76719979e-03 2.90116407e-07 3.10952455e-03 5.70037953e-06
%  1.33549520e-05]

% baseline
% log rse [ 3.94008568 -2.49808431 -3.06113587 -2.06672228 -2.30289369]
% log 1-ACC [-3.33990332 -2.80013793 -3.36229269 -2.84158398 -2.80835118]

%best
% log rse [ 3.11275337 -3.14340423 -3.98422582 -2.92144808 -3.01291351]
% log 1-ACC [-3.21066235 -3.55167183 -4.28609952 -3.25368933 -3.34871348]

% \begin{table}[]
%     \centering
%     \caption{The MSE and ACC scores of the baseline model the model with the optimal configuration from the search results.}
%     \begin{tabular}{lcccc}
%     \toprule
%                            & \multicolumn{2}{c}{Baseline} & \multicolumn{2}{c}{Optimal}\\
%     \midrule
%                            & MSE                          & ACC      & MSE & ACC \\
%     \midrule
%     Layer Thickness        &  $6.262e-2 $ &  $0.999$   & $9.767e-3$ & $0.999$      \\
%     Salinity               &  $1.282e-6$  &  $0.998$   & $2.901e-7$ & $0.999$    \\ 
%     Temperature            &  $2.604e-2$  &  $0.999$   & $3.109e-3$ & $0.999$         \\
%     Meridional V.          &  $4.079e-6$  &  $0.998$   & $5.703e-6$ & $0.999$      \\ 
%     Zonal V.               &  $6.849e-5$  &  $0.998$   & $1.335e-5$ & $0.999$   \\ 
%     \bottomrule
%     \end{tabular}
% \end{table}

% \begin{figure}
%     \centering
%     % \begin{subfigure}{.49\textwidth}
%     % \centering
%     % \includegraphics[width=\linewidth]{figs/2024-01-10_scatter_PF.pdf}
%     % \caption{Caption}
%     % \label{fig:enter-label}
%     % \end{subfigure}
%     %
%     \begin{subfigure}{.49\textwidth}
%     \centering
%     \includegraphics[width=\linewidth]{figs/defaultBestTrainCurves.pdf}
%     \caption{Caption}
%     \label{fig:enter-label}
%     \end{subfigure}
%     %
%     \begin{subfigure}{.49\textwidth}
%         \centering
%         \includegraphics[width=\linewidth]{figs/rollout_compare.pdf}
%         \caption{Caption \todo[inline]{train default and best for longer time about 100 epochs and compare}}
%         \label{fig:enter-label}
%     \end{subfigure}
% \end{figure}

%\todo{Add a rollout comparison table.}

\section{Conlusion}\label{sec:conclusion}
We explored streamlining data-driven ocean modeling using Fourier neural operators with hyperparameter search. In particular, we focused on forecasting four prognostic variables for an idealized baroclinic wind-driven ocean model. We proposed incorporating the anomaly correlation coefficient in the loss function as a simple improvement to address the potential drawbacks of the mean squared error loss function in model training. The hyperparameter search adopted multiobjective optimization to minimize validation MSE and maximize ACC simultaneously. The optimal configuration obtained from the search results improves the model performance for all four variables, which leads to immense improvement in the model's autoregressive rollout performance. We have demonstrated that hyperparameter optimization can dramatically improve the performance of FNOs by targeting specific objectives, which consolidates high-performing ocean modeling using deep neural networks such as FNOs. Future work includes exploring more efficient function evaluation strategies during the search that allow more reduction in resource consumption, such as one-epoch and multifidelity approaches, and analyzing the effect of negative ACC on the loss landscape and optimization.

\paragraph{Acknowledgments}This research used resources from the Argonne Leadership Computing Facility at Argonne National Laboratory. This material is based upon work supported by the U.S. Department of Energy, Office of Science, Office of Advanced Scientific Computing Research, and Office of Biological and Environmental Research, Scientific Discovery through Advanced Computing (SciDAC) program under Award Number(s) 9233218CNA000001.

% \conflictsofinterest{Declare conflicts of interest or state ``The authors declare no conflict of interest.'' Authors must identify and declare any personal circumstances or interest that may be perceived as inappropriately influencing the representation or interpretation of reported research results. Any role of the funders in the design of the study; in the collection, analyses or interpretation of data; in the writing of the manuscript; or in the decision to publish the results must be declared in this section. If there is no role, please state ``The funders had no role in the design of the study; in the collection, analyses, or interpretation of data; in the writing of the manuscript; or in the decision to publish the~results''.} 

%\newpage
% \reftitle{References}
\bibliographystyle{ieeetr}

% Please provide either the correct journal abbreviation (e.g. according to the “List of Title Word Abbreviations” http://www.issn.org/services/online-services/access-to-the-ltwa/) or the full name of the journal.
% Citations and References in Supplementary files are permitted provided that they also appear in the reference list here. 

%=====================================
% References, variant A: external bibliography
%=====================================
\bibliography{ref}

%=====================================
% References, variant B: internal bibliography
%=====================================

% If authors have biography, please use the format below
%\section*{Short Biography of Authors}
%\bio
%{\raisebox{-0.35cm}{\includegraphics[width=3.5cm,height=5.3cm,clip,keepaspectratio]{Definitions/author1.pdf}}}
%{\textbf{Firstname Lastname} Biography of first author}
%
%\bio
%{\raisebox{-0.35cm}{\includegraphics[width=3.5cm,height=5.3cm,clip,keepaspectratio]{Definitions/author2.jpg}}}
%{\textbf{Firstname Lastname} Biography of second author}

% For the MDPI journals use author-date citation, please follow the formatting guidelines on http://www.mdpi.com/authors/references
% To cite two works by the same author: \citeauthor{ref-journal-1a} (\citeyear{ref-journal-1a}, \citeyear{ref-journal-1b}). This produces: Whittaker (1967, 1975)
% To cite two works by the same author with specific pages: \citeauthor{ref-journal-3a} (\citeyear{ref-journal-3a}, p. 328; \citeyear{ref-journal-3b}, p.475). This produces: Wong (1999, p. 328; 2000, p. 475)

%%%%%%%%%%%%%%%%%%%%%%%%%%%%%%%%%%%%%%%%%%
%% for journal Sci
%\reviewreports{\\
%Reviewer 1 comments and authors’ response\\
%Reviewer 2 comments and authors’ response\\
%Reviewer 3 comments and authors’ response
%}
%%%%%%%%%%%%%%%%%%%%%%%%%%%%%%%%%%%%%%%%%%

\end{document}